\documentclass[10pt,twocolumn,letterpaper]{article}

\usepackage{cvpr}              

\usepackage{graphicx}
\usepackage{amsmath}
\usepackage{amssymb}
\usepackage{booktabs}

\usepackage[pagebackref,breaklinks,colorlinks]{hyperref}

\usepackage[capitalize]{cleveref}
\crefname{section}{Sec.}{Secs.}
\Crefname{section}{Section}{Sections}
\Crefname{table}{Table}{Tables}
\crefname{table}{Tab.}{Tabs.}

\def\cvprPaperID{6484} 
\def\confName{CVPR}
\def\confYear{2022}

\begin{document}

\title{NLX-GPT: A Model for Natural Language Explanations in Vision and Vision-Language Tasks}

\author{Fawaz Sammani$^{1}$, Tanmoy Mukherjee$^{1,2}$, Nikos Deligiannis$^{1,2}$\\
$^{1}$ETRO Department, Vrije Universiteit Brussel, Pleinlaan 2, B-1050 Brussels, Belgium
\\$^{2}$imec, Kapeldreef 75, B-3001 Leuven, Belgium\\
{\tt\small fawaz.sammani@vub.be, \tt\small tmukherj@etrovub.be, \tt\small ndeligia@etrovub.be
}
}

\maketitle

\begin{abstract}
Natural language explanation (NLE) models aim at explaining the decision-making process of a black box system via generating natural language sentences which are human-friendly, high-level and fine-grained. Current NLE models\footnote{Throughout this paper, we refer to NLE models as Natural Language Explanation models aimed for vision and vision-language tasks.} explain the decision-making process of a vision or vision-language model (a.k.a., task model), e.g., a VQA model, via a language model (a.k.a., explanation model), e.g., GPT. Other than the additional memory resources and inference time required by the task model, the task and explanation models are completely independent, which disassociates the explanation from the reasoning process made to predict the answer. We introduce NLX-GPT, a general, compact and faithful language model that can simultaneously predict an answer and explain it. We first conduct pre-training on large scale data of image-caption pairs for general understanding of images, and then formulate the answer as a text prediction task along with the explanation. Without region proposals nor a task model, our resulting overall framework attains better evaluation scores, contains much less parameters and is 15$\times$ faster than the current SoA model. We then address the problem of evaluating the explanations which can be in many times generic, data-biased and can come in several forms. We therefore design 2 new evaluation measures: (1) explain-predict and (2) retrieval-based attack, a self-evaluation framework that requires no labels. Code is at: \href{https://github.com/fawazsammani/nlxgpt}{https://github.com/fawazsammani/nlxgpt}. 

\end{abstract}

\begin{figure}
    \centering
    \includegraphics[width=0.53\textwidth]{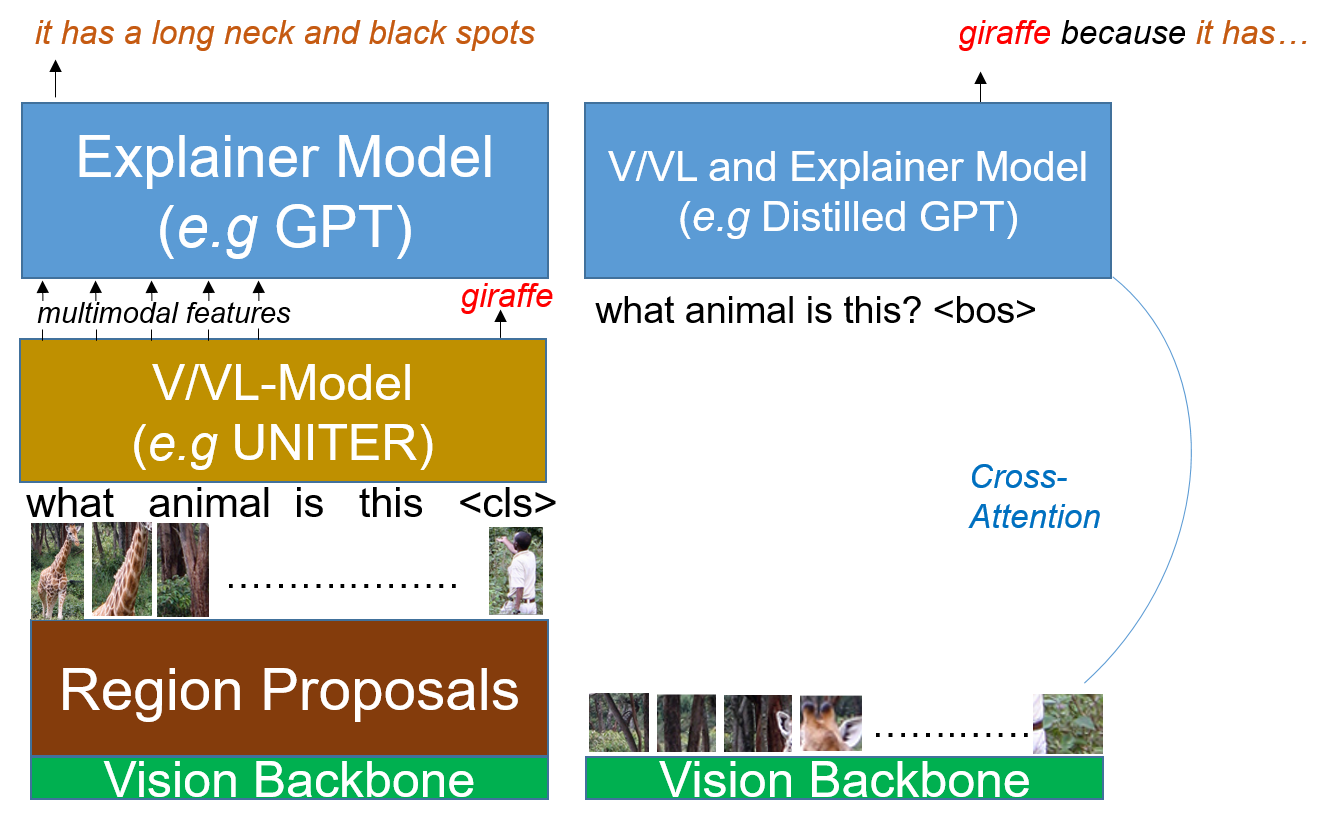}
    \caption{A comparison between previous models (left) and ours (right). Our model solely requires a visual encoder and a language model. We model the answer as a text prediction task along with the explanation. Best viewed in color. }
    \label{demo}
\end{figure}

\section{Introduction}

Deep learning models have enabled extraordinary breakthroughs in a variety of vision tasks (such as image classification \cite{Krizhevsky2012ImageNetCW, Dosovitskiy2020AnII, He2016DeepRL}) and vision-language tasks (such as visual question answering \cite{Agrawal2015VQAVQ, Anderson2018BottomUpAT, Yu2019DeepMC}, visual entailment \cite{Xie2019VisualEA}, image captioning \cite{Vinyals2015ShowAT, Pan2020XLinearAN, Cornia2020MeshedMemoryTF, Sammani2020ShowEA}, and more), achieving promising performance. However, they are black box systems. For these models to be deployed in everyday life, explaining their decision-making process becomes critical for several reasons such as trust, accountability, and model bias understanding and correctness. Different from visual or textual explanations which highlight regions or tokens in an image or sentence that lead to a specific prediction \cite{Selvaraju2019GradCAMVE, Simonyan2014DeepIC, Sundararajan2017AxiomaticAF, Bach2015OnPE}, natural language explanation models \cite{Camburu2018eSNLINL, Narang2020WT5TT} explain the decision-making process of a model through natural language sentences. These sentences are easy to understand by humans and are much more detailed than highlighted regions or tokens. Recently, NLE for vision and vision-language (VL) tasks has been introduced \cite{Park2018MultimodalEJ, Wu2019FaithfulME, Marasovi2020NaturalLR, Kayser2021eViLAD}. In this work, we focus on explaining models aimed for vision and vision-language tasks. Current NLE models \cite{Park2018MultimodalEJ, Wu2019FaithfulME, Marasovi2020NaturalLR, Kayser2021eViLAD} first utilize a VL-model to get an answer for the task at hand (e.g., a visual question answering (VQA) model). The outputs of the VL-model (answer and multimodal features) along with the question are then fed to language model (e.g., LSTM or Transformer) to get an explanation for the answer (see Figure~\ref{demo}). At training time, the language model is trained to produce explanations for the ground-truth answers with a NLE dataset. At test time, the output of a VL-model is utilized to predict an answer which is fed to the language model to get the explanation. This paradigm has two disadvantages. First, the addition of the task model requires higher storage and memory requirements (typically, approx. 80M and 300M parameters for small and large models, respectively). Second, the VL-model and language model are completely independent of each other, which disconnects the explanations from the reasoning process made to predict the answer. Finally, automatic NLE measures that evaluate the generated explanation do not always reflect the correctness, reasoning and semantic meaning of the explanations since explanations can come in different forms and may learn correlations and bias in the dataset.

In summary, we make the following contributions: 
\begin{itemize}
\item We propose NLX-GPT, a model which can simultaneously predict an answer and explain it, by formulating the answer prediction as a text generation task \textit{along with} the explanation. This eliminates the need for a VL-model to provide an answer and associates the explanation with the reasoning process made to predict the answer.
\item Our method outperforms previous works on most metrics while being 15$\times$ faster and requiring less memory resources. We further present an ablation analysis of the steps and components of our model, demonstrating that each aspect contributes non-trivially to the final performance of the model. 
\item We present two new evaluation frameworks for NLE which can reflect the correctness, reasoning, semantic meaning and the degree of biasness of the generated explanations
\end{itemize}

\section{Related Work}
The first work on NLE was proposed by \cite{Hendricks2016GeneratingVE} for vision tasks. It was then extended to video tasks \cite{Kim2018TextualEF}, vision-language tasks \cite{Park2018MultimodalEJ, Li2018VQAEEE, Wu2019FaithfulME, Marasovi2020NaturalLR, Kayser2021eViLAD} and NLP tasks \cite{Narang2020WT5TT, Camburu2018eSNLINL, Jang2021AreTR, Rajani2019ExplainYL}. The authors of \cite{Hendricks2016GeneratingVE} propose a discriminative objective that can take into account class-discriminative image properties which justify a classification prediction. Later, \cite{Park2018MultimodalEJ} proposed two datasets: ACT-X and VQA-X. The former is used to explain decisions of activity recognition models, while the later is used to explain decisions of VQA models. The authors used the MCB VQA model \cite{Fukui2016MultimodalCB} as the task model and an LSTM as the explanation model. The authors of \cite{Wu2019FaithfulME} focus on generating more faithful multimodal explanations and filters out the human textual explanations whose Grad-CAM visual explanation does not align with the Grad-CAM of the predicted answer. Their task model is the Up-Down VQA model \cite{Anderson2018BottomUpAT} and their explanation model is the Up-Down LSTM model \cite{Anderson2018BottomUpAT}. In \cite{Li2018VQAEEE} , an automatic dataset of explanations for the VQA task is constructed, in which the explanations are essentially image captions describing the image. \cite{Marasovi2020NaturalLR} uses different vision models to encode the image and then inputs them along with the ground-truth answer and question to a GPT-2 model \cite{Radford2019LanguageMA} to generate an explanation. Finally, \cite{Kayser2021eViLAD} corrects the eSNLI-VE dataset \cite{Do2020eSNLIVE20CV} for explanations of the visual entailment task. Their e-UG model is composed of UNITER \cite{Chen2020UNITERUI} as the task model and GPT-2 \cite{Radford2019LanguageMA} as the explanation model. 

As observed, all these works rely on a task model, which brings the disadvantages discussed previously. Our proposed NLX-GPT tackles these problems by eliminating the task model and produces the answer as a text generation task along with the explanation. 

\section{NLX-GPT}

\begin{figure*}
    \centering
    \includegraphics[width=\textwidth]{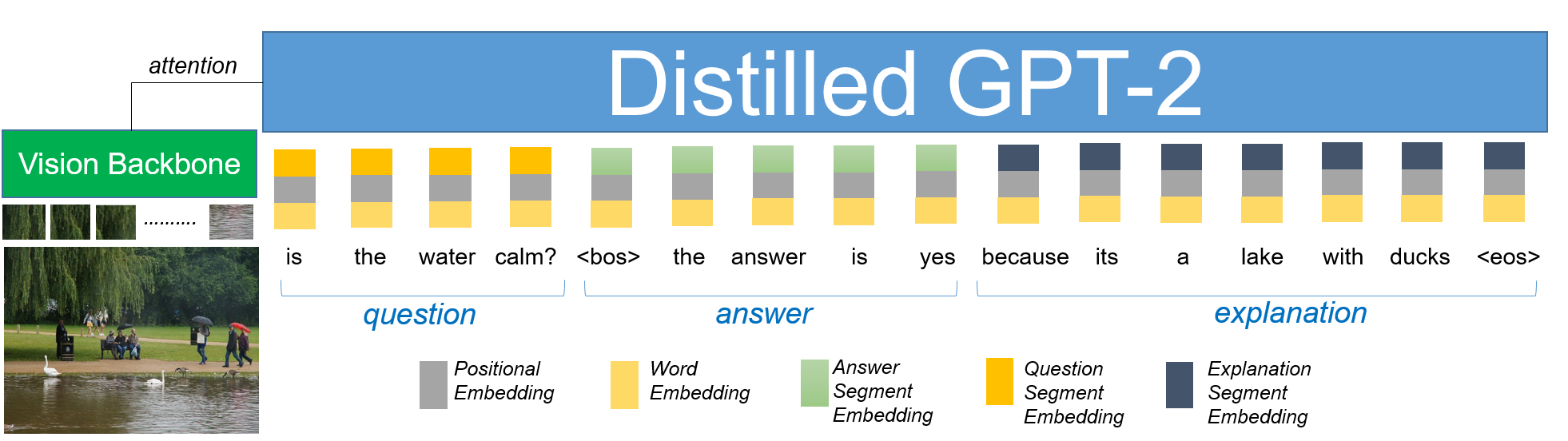}
    \caption{A schema of the proposed NLX-GPT model. At test time, we supply the question and {\fontfamily{qcr}\selectfont <bos>} token, and start generating the answer and explanation}
    \label{model}
\end{figure*}

Consider a task model $M_{T}$ which performs a specific vision or vision-language task. For example, $M_{T}$ can be an answering model for VQA, visual entailment or activity recognition. Also, consider an explainer model $M_{E}$ which provides an explanation for the output of the vision-language model. Previous models \cite{Park2018MultimodalEJ, Wu2019FaithfulME, Marasovi2020NaturalLR, Kayser2021eViLAD} stack both $M_{T}$ and $M_{E}$ on top of each other. That is, $M_{T}$ first answers the task, and $M_{E}$ provides an explanation for the answer. Drawing inspiration from~\cite{Rajani2019ExplainYL, Cho2021UnifyingVT}, in our work we eliminate $M_{T}$ and allow $M_{E}$ to address both objectives. In other words, $M_{E}$ simultaneously predicts an answer and explains it. This is achieved by formulating the answer prediction as a text generation task \textit{along with} the explanation. This has two advantages: First, it eliminates the high memory requirements of $M_{T}$, which can reach to over 300M parameters \cite{Chen2020UNITERUI, Li2020OscarOA}, and reduces the inference time. Second, unlike previous models, where $M_{T}$ and $M_{E}$ are completely independent, the generated answer and explanation from our model become associated with each other, in the sense that the explanation is intrinsic, internally affiliated and connected to the reasoning process made to predict the answer. In fact, the answering task may perform even better than having a separate model $M_{T}$. For example, our VQA accuracy outperforms the SoA \cite{Kayser2021eViLAD} which uses UNITER \cite{Chen2020UNITERUI} as $M_{T}$ by 3\%. This shows the strong ability of modeling the answering task as a text-generation task \textit{along with}  its explanation and vice-versa. There are other reasons for why removing a VL-model is advantageous. We refer readers to Section 7 of the Appendix for more information. In our case, we choose $M_{T}$ to be a distilled version \cite{Sanh2019DistilBERTAD} of the GPT-2 transformer language model \cite{Brown2020LanguageMA}. Our model is an encoder-decoder architecture, as originally proposed in \cite{Vaswani2017AttentionIA}. The encoder is a visual backbone that encodes the image and the decoder is the distilled GPT-2. We consider all sub-inputs (question/hypothesis, answer and explanation) as a single sequence to $M_{E}$ and train $M_{E}$ with the cross-entropy objective to generate a sequence $\boldsymbol{w} = w_1, w_2, \dots, w_T$~of~$T$~words (containing both the answer and explanation) by minimizing the negative log-likelihood:
\begin{equation}
\label{eq:crossEntropyLoss}
\mathcal{L}=- \sum_{t=1}^{T} \log p_{\theta}\left(w_{t} \mid \boldsymbol{w}_{<t}\right),
\end{equation}
where $\boldsymbol{w}_{<t}$ denotes the words before word $t$. A schema of our model is shown in Figure~\ref{model}. Note that our model can still be used to explain the answer of any external VL-model by simply appending the answer output of that VL-model after the question at the input to our NLX-GPT, which conditions the explanation on that answer. In fact, this becomes a special case of our general model. 

At the same time, we aim to deviate away from visual encoders biased towards a specific task (e.g., image classification) as well as from time-expensive bottom-up features \cite{Anderson2018BottomUpAT}. We would instead like to fully rely on grid features. To this, we utilize the CLIP vision encoder \cite{Radford2021LearningTV} since the visual features are (1) non-biased and general and (2) close to the language embedding space of GPT-2 due to the contrastive learning objective of CLIP, which makes the fusing of visual and linguistic information easier. Without region proposals nor a VL-model, our model becomes significantly faster and more memory efficient than previous models, as shown in Table 1. However, we still demonstrate in later sections that even with a ResNet-101 visual encoder pretrained on ImageNet-1K, our model can still outperform previous models that also use a ResNet-101, even without bottom-up object level features \cite{Anderson2018BottomUpAT}. With bottom-up object-level features, our model significantly outperforms the SoA even without using an additional BERT-based multimodal feature extractor (such as UNITER \cite{Chen2020UNITERUI}). In the next subsections, we will explain the two stages we conduct on our model: pretraining and finetuning. Please note that our visual encoder is fixed and not fine-tuned at any time during pretraining or finetuning.

\begin{table*}[]
\caption{Comparison between other models in terms of the visual encoder, region proposals, VL-model, explanation generator, inference time and total number of parameters. Note that VL-models in previous works act as multi-modal feature extractors and answering models.}
\scalebox{0.95}{
\begin{tabular}{|c|cccccc|}
\hline
Model          & Visual   Encoder & Region   Proposals & VL-model     & Explanation   Generator & Time (ms) & Parameters   (M) \\ \hline
FME            & ResNet-101       & $\checkmark$                  & Up-Down VQA & Up-Down   LSTM          & $\sim$910       & 142             \\ 
RVT            & ResNet-101       & $\checkmark$                   & BERT           & GPT-2                   & $\sim$925       & 277             \\ 
e-UG           & ResNet-101       & $\checkmark$                  & UNITER         & GPT-2                   & $\sim$929       & 277              \\  \hline
NLX-GPT     & ResNet-101       & $\times$                  & $\times$              & Distilled   GPT-2       & $\sim$93       & 138             \\
NLX-GPT  & ViT              & $\times$                  & $\times$              & Distilled   GPT-2       & $\sim$55        & 182              \\ \hline
\end{tabular}}
\end{table*}

\begin{table*}[]
\centering
\caption{Unfiltered Scores for VQA-X and ACT-X. B4, M R, C, S are short for: BLEU-4, METEOR, ROUGE-L, CIDER and SPICE. Unfiltered scores on e-SNLI-VE are 11.9, 18.2, 32.5, 1.09 and 33.0, respectively. }
\begin{tabular}{|c|ccccc|cccccc|}
\hline
                 & \multicolumn{5}{c|}{\textbf{VQA-X}}                                                                                                                               & \multicolumn{6}{c|}{\textbf{ACT-X}}                                                                                                                                                                    \\ \hline
Approach         & \multicolumn{1}{c|}{B4}            & \multicolumn{1}{c|}{M}             & \multicolumn{1}{c|}{R}           & \multicolumn{1}{c|}{C}             & S             & \multicolumn{1}{c|}{B4}            & \multicolumn{1}{c|}{M}             & \multicolumn{1}{c|}{R}             & \multicolumn{1}{c|}{C}             & \multicolumn{1}{c|}{S}             & Human         \\ \hline
CAPS\cite{Park2018MultimodalEJ}             & \multicolumn{1}{c|}{5.9}           & \multicolumn{1}{c|}{12.6}          & \multicolumn{1}{c|}{26.3}          & \multicolumn{1}{c|}{35.2}          & 11.9          & \multicolumn{1}{c|}{5.2}           & \multicolumn{1}{c|}{11.0}          & \multicolumn{1}{c|}{26.5}          & \multicolumn{1}{c|}{10.4}          & \multicolumn{1}{c|}{4.6}           & 22.9          \\ 
PJ-X  \cite{Park2018MultimodalEJ}           & \multicolumn{1}{c|}{19.5}          & \multicolumn{1}{c|}{18.2}          & \multicolumn{1}{c|}{43.4}          & \multicolumn{1}{c|}{71.3}          & 15.1          & \multicolumn{1}{c|}{15.3}          & \multicolumn{1}{c|}{15.6}          & \multicolumn{1}{c|}{40.0}          & \multicolumn{1}{c|}{22.0}          & \multicolumn{1}{c|}{7.2}           & 38.2          \\ 
FME   \cite{Wu2019FaithfulME}           & \multicolumn{1}{c|}{24.4}          & \multicolumn{1}{c|}{19.5}          & \multicolumn{1}{c|}{47.4}          & \multicolumn{1}{c|}{88.8}          & 17.9          & \multicolumn{1}{c|}{-}             & \multicolumn{1}{c|}{-}             & \multicolumn{1}{c|}{-}             & \multicolumn{1}{c|}{-}             & \multicolumn{1}{c|}{-}             & -             \\  \hline
NLX-GPT   (w/o pretraining) & \multicolumn{1}{c|}{23.8}          & \multicolumn{1}{c|}{20.3}          & \multicolumn{1}{c|}{47.2}          & \multicolumn{1}{c|}{89.2}          & 18.3          & \multicolumn{1}{c|}{25.6}          & \multicolumn{1}{c|}{21.4}          & \multicolumn{1}{c|}{48.0}          & \multicolumn{1}{c|}{63.5}          & \multicolumn{1}{c|}{15.4}          & -             \\ 
NLX-GPT   (w/ pretraining) & \multicolumn{1}{c|}{\textbf{25.6}} & \multicolumn{1}{c|}{\textbf{21.5}} & \multicolumn{1}{c|}{\textbf{48.7}} & \multicolumn{1}{c|}{\textbf{97.2}} & \textbf{20.2} & \multicolumn{1}{c|}{\textbf{28.1}} & \multicolumn{1}{c|}{\textbf{22.6}} & \multicolumn{1}{c|}{\textbf{49.7}} & \multicolumn{1}{c|}{\textbf{74.9}} & \multicolumn{1}{c|}{\textbf{17.6}} & \textbf{89.0} \\ \hline
\end{tabular}
\end{table*}

\begin{table*}[]
\centering
\caption{Filtered scores for VQA-X and e-SNLI-VE. BS stands for BERTScore. $^{\dagger}$ Since the CIDER score is exceptionally high, we suspect a difference in the CIDER idf weights used by the e-VIL \cite{Kayser2021eViLAD} authors. We therefore report our scores with a different evaluation API \cite{sharma2017nlgeval}. Our resulting scores on this API on NLX-GPT (w/ concepts) for B1, B4, R-L, M and C are: 35.3, 11.9, 33.1, 18.7, 112.5}
\begin{tabular}{|cccccccccccc|}
\hline
\multicolumn{12}{|c|}{VQA-X}                                                                                                                                                                           \\ \hline
\multicolumn{1}{|c|}{}                    & \multicolumn{1}{c|}{B1}            & \multicolumn{1}{c|}{B2}            & \multicolumn{1}{c|}{B3}            & \multicolumn{1}{c|}{B4}            & \multicolumn{1}{c|}{R-L}           & \multicolumn{1}{c|}{M}             & \multicolumn{1}{c|}{C}               & \multicolumn{1}{c|}{S}             & \multicolumn{1}{c|}{BS}     & \multicolumn{1}{c|}{Human}    & Task Acc. \\ \hline
\multicolumn{1}{|c|}{PJ-X \cite{Park2018MultimodalEJ}}                & \multicolumn{1}{c|}{57.4}          & \multicolumn{1}{c|}{42.4}          & \multicolumn{1}{c|}{30.9}          & \multicolumn{1}{c|}{22.7}          & \multicolumn{1}{c|}{46.0}          & \multicolumn{1}{c|}{19.7}          & \multicolumn{1}{c|}{82.7}            & \multicolumn{1}{c|}{17.1}          & \multicolumn{1}{c|}{84.6}          & \multicolumn{1}{c|}{65.4}     & 76.4     \\ 
\multicolumn{1}{|c|}{FME \cite{Wu2019FaithfulME}}                 & \multicolumn{1}{c|}{59.1}          & \multicolumn{1}{c|}{43.4}          & \multicolumn{1}{c|}{31.7}          & \multicolumn{1}{c|}{23.1}          & \multicolumn{1}{c|}{47.1}          & \multicolumn{1}{c|}{20.4}          & \multicolumn{1}{c|}{87.0}            & \multicolumn{1}{c|}{18.4}          & \multicolumn{1}{c|}{85.2}          & \multicolumn{1}{c|}{63.2}    & 75.5       \\ 
\multicolumn{1}{|c|}{RVT \cite{Marasovi2020NaturalLR}}                 & \multicolumn{1}{c|}{51.9}          & \multicolumn{1}{c|}{37.0}          & \multicolumn{1}{c|}{25.6}          & \multicolumn{1}{c|}{17.4}          & \multicolumn{1}{c|}{42.1}          & \multicolumn{1}{c|}{19.2}          & \multicolumn{1}{c|}{52.5}            & \multicolumn{1}{c|}{15.8}          & \multicolumn{1}{c|}{85.7}          & \multicolumn{1}{c|}{67.1}     & 68.6      \\ 
\multicolumn{1}{|c|}{QA-only \cite{Kayser2021eViLAD}}             & \multicolumn{1}{c|}{51.0}          & \multicolumn{1}{c|}{36.4}          & \multicolumn{1}{c|}{25.3}          & \multicolumn{1}{c|}{17.3}          & \multicolumn{1}{c|}{41.9}          & \multicolumn{1}{c|}{18.6}          & \multicolumn{1}{c|}{49.9}            & \multicolumn{1}{c|}{14.9}          & \multicolumn{1}{c|}{85.3}          & \multicolumn{1}{c|}{-}          &--    \\ 
\multicolumn{1}{|c|}{e-UG \cite{Kayser2021eViLAD}}                & \multicolumn{1}{c|}{57.3}          & \multicolumn{1}{c|}{42.7}          & \multicolumn{1}{c|}{31.4}          & \multicolumn{1}{c|}{23.2}          & \multicolumn{1}{c|}{45.7}          & \multicolumn{1}{c|}{22.1}          & \multicolumn{1}{c|}{74.1}            & \multicolumn{1}{c|}{20.1}          & \multicolumn{1}{c|}{\textbf{87.0}} & \multicolumn{1}{c|}{71.5}     & 80.5      \\ \hline
\multicolumn{1}{|c|}{NLX-GPT}              & \multicolumn{1}{c|}{\textbf{64.2}} & \multicolumn{1}{c|}{\textbf{49.5}} & \multicolumn{1}{c|}{\textbf{37.6}} & \multicolumn{1}{c|}{\textbf{28.5}} & \multicolumn{1}{c|}{\textbf{51.5}} & \multicolumn{1}{c|}{\textbf{23.1}} & \multicolumn{1}{c|}{\textbf{110.6}}  & \multicolumn{1}{c|}{\textbf{22.1}} & \multicolumn{1}{c|}{86.9}          & \multicolumn{1}{c|}{\textbf{83.22}} & \multicolumn{1}{c|}{\textbf{83.07}} \\ \hline
\multicolumn{12}{|c|}{e-SNLI-VE}  \\ \hline
\multicolumn{1}{|c|}{}                    & \multicolumn{1}{c|}{B1}            & \multicolumn{1}{c|}{B2}            & \multicolumn{1}{c|}{B3}            & \multicolumn{1}{c|}{B4}            & \multicolumn{1}{c|}{R-L}           & \multicolumn{1}{c|}{M}             & \multicolumn{1}{c|}{C}               & \multicolumn{1}{c|}{S}             & \multicolumn{1}{c|}{BS}     & \multicolumn{1}{c|}{Human}   & Task Acc.       \\ \hline
\multicolumn{1}{|c|}{PJ-X \cite{Park2018MultimodalEJ}}                & \multicolumn{1}{c|}{29.4}          & \multicolumn{1}{c|}{18.0}          & \multicolumn{1}{c|}{11.3}          & \multicolumn{1}{c|}{7.3}           & \multicolumn{1}{c|}{28.6}          & \multicolumn{1}{c|}{14.7}          & \multicolumn{1}{c|}{72.5}            & \multicolumn{1}{c|}{24.3}          & \multicolumn{1}{c|}{79.1}          & \multicolumn{1}{c|}{59.6}        &  69.2   \\ 
\multicolumn{1}{|c|}{FME \cite{Wu2019FaithfulME}}                 & \multicolumn{1}{c|}{30.6}          & \multicolumn{1}{c|}{19.2}          & \multicolumn{1}{c|}{12.4}          & \multicolumn{1}{c|}{8.2}           & \multicolumn{1}{c|}{29.9}          & \multicolumn{1}{c|}{15.6}          & \multicolumn{1}{c|}{83.6}            & \multicolumn{1}{c|}{26.8}          & \multicolumn{1}{c|}{79.7}          & \multicolumn{1}{c|}{58.5}       & 73.7    \\ 
\multicolumn{1}{|c|}{RVT \cite{Marasovi2020NaturalLR}}                 & \multicolumn{1}{c|}{29.9}          & \multicolumn{1}{c|}{19.8}          & \multicolumn{1}{c|}{13.6}          & \multicolumn{1}{c|}{9.6}           & \multicolumn{1}{c|}{27.3}          & \multicolumn{1}{c|}{18.8}          & \multicolumn{1}{c|}{81.7}            & \multicolumn{1}{c|}{32.5}          & \multicolumn{1}{c|}{81.1}          & \multicolumn{1}{c|}{59.4}      & 72.0     \\ 
\multicolumn{1}{|c|}{QA-only \cite{Kayser2021eViLAD}}             & \multicolumn{1}{c|}{29.8}          & \multicolumn{1}{c|}{19.7}          & \multicolumn{1}{c|}{13.5}          & \multicolumn{1}{c|}{9.5}           & \multicolumn{1}{c|}{27.0}          & \multicolumn{1}{c|}{18.7}          & \multicolumn{1}{c|}{80.4}            & \multicolumn{1}{c|}{32.1}          & \multicolumn{1}{c|}{81.1}          & \multicolumn{1}{c|}{-}        &--      \\ 
\multicolumn{1}{|c|}{e-UG \cite{Kayser2021eViLAD}}                & \multicolumn{1}{c|}{30.1}          & \multicolumn{1}{c|}{19.9}          & \multicolumn{1}{c|}{13.7}          & \multicolumn{1}{c|}{9.6}           & \multicolumn{1}{c|}{27.8}          & \multicolumn{1}{c|}{\textbf{19.6}} & \multicolumn{1}{c|}{85.9}            & \multicolumn{1}{c|}{\textbf{34.5}} & \multicolumn{1}{c|}{\textbf{81.7}} & \multicolumn{1}{c|}{\textbf{68.9}} & \textbf{79.5} \\ \hline
\multicolumn{1}{|c|}{NLX-GPT   (w/o Concepts)} & \multicolumn{1}{c|}{35.7}          & \multicolumn{1}{c|}{24.0}          & \multicolumn{1}{c|}{16.8}          & \multicolumn{1}{c|}{11.9}          & \multicolumn{1}{c|}{33.4}          & \multicolumn{1}{c|}{18.1}          & \multicolumn{1}{c|}{114.7}           & \multicolumn{1}{c|}{32.1}          & \multicolumn{1}{c|}{80.6}          & \multicolumn{1}{c|}{66.3}    & --      \\ 
\multicolumn{1}{|c|}{NLX-GPT   (w/ Concepts)}  & \multicolumn{1}{c|}{\textbf{37.0}} & \multicolumn{1}{c|}{\textbf{25.3}} & \multicolumn{1}{c|}{\textbf{17.9}} & \multicolumn{1}{c|}{\textbf{12.9}} & \multicolumn{1}{c|}{\textbf{34.2}} & \multicolumn{1}{c|}{18.8}          & \multicolumn{1}{c|}{\textbf{117.4$^{\dagger}$}} & \multicolumn{1}{c|}{33.6}          & \multicolumn{1}{c|}{80.8}          & \multicolumn{1}{c|}{67.4}      & 73.91     \\ \hline
\end{tabular}
\end{table*}

\subsection{Pretraining}
Training an NLE model to explain the decision-making process of an answer given a particular image requires strong image understanding in the first place. Otherwise, the model may be susceptible to learning correlations and bias in the dataset, or overfitting due to the small-scale NLE dataset. In later sections, we make this evident through visualization. It is also shown in \cite{Desai2021VirTexLV} how image understanding greatly helps image classification and object detection tasks. Following the trend of vision-language pretraining \cite{Zhou2020UnifiedVP, Chen2020UNITERUI, Li2020OscarOA, Kim2021ViLTVT}, we pretrain our Distilled GPT-2 on a large-scale corpus of image-caption pairs. We choose image captioning as our pre-training task because (1) it is aligned with our downstream task of text generation and (2) image captions provide comprehensive image understanding by describing objects, their attributes and their relationships. Particularly, we use data from COCO captions  \cite{Lin2014MicrosoftCC}, Flickr30k \cite{Plummer2015Flickr30kEC}, visual genome (VG) \cite{Krishna2016VisualGC} and image paragraph captioning \cite{Krause2017AHA}. In the case of VG region descriptions, we combine region descriptions per image to form a paragraph.  More details can be found in the Appendix. We use the cross-entropy objective loss (as in~\eqref{eq:crossEntropyLoss}) to train our model to generate a caption $\boldsymbol{c}$ describing an image $I$ in an autoregressive manner. Other pre-training objectives such as image feature regression, masked language modelling and image-text matching can be used; however, we find that the results are already satisfactory with cross-entropy training. It is also shown in \cite{Xia2021XGPTCG} that these additional objectives only contribute a little to the overall performance. We pretrain the model with a batch size of 768; we refer to the Appendix for further implementation details.

\subsection{Concept Detection}
In the e-SNLI-VE dataset~\cite{Kayser2021eViLAD}, we find that pre-training does not help our model generalize. We therefore propose to compensate for image understanding by predicting image concepts with a simple multilayer perceptron (MLP) layer attached on top of the visual encoder and trained on the Visual Genome dataset \cite{Krishna2016VisualGC}. Particularly, we use binary cross-entropy as the loss function to train an $N$-way multi-label classifier, where $N$ is the concept vocabulary:
\begin{equation}
\mathcal{L}_{\mathrm{P}}=-\sum_{i\in \mathcal{B}} \sum_{j=1}^{N} p_{i j} \log \hat{p}_{i j}+\left(1-p_{i j}\right) \log \left(1-\hat{p}_{i j}\right), 
\end{equation}
where $p_{i j}$ is the $j$-th target for the $i$-th datapoint in the batch~$\mathcal{B}$  and $\hat{p}_{i j}$ is the sigmoid probability output. After detecting the concepts, we append them to the GPT-2 input and model a conditional language generation task \cite{Keskar2019CTRLAC}. Further implementation details can be found in the Appendix. 

\subsection{Finetuning}
After the pretraining stage, we finetine our NLX-GPT model on NLE tasks using a batch size of 32. Particularly, we choose NLE for visual-question answering, visual entailment and visual commonsense reasoning (VQA-X~\cite{Park2018MultimodalEJ}, e-SNLI-VE~\cite{Kayser2021eViLAD} and VCR~\cite{Zellers2019FromRT}) as vision-language tasks, and NLE for activity recognition (ACT-X)~\cite{Park2018MultimodalEJ} as the vision task. See Section 4 for more details about these datasets. 

\subsection{Evaluation Measures}

\begin{figure*}
    \centering
    \includegraphics[width=\textwidth]{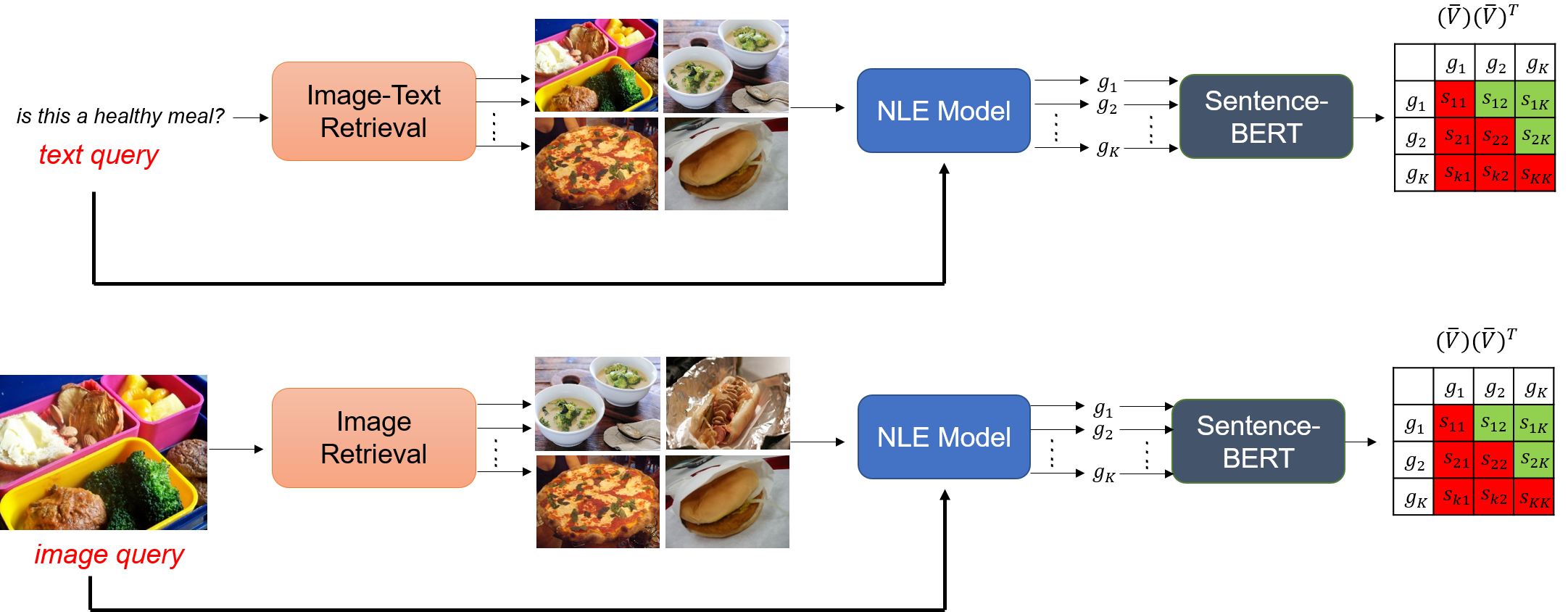}
    \caption{The retrieval-based attack evaluation framework. On top, we show how to measure the biasness to text in vision-language tasks. At the bottom, we show the process for vision tasks. }
    \label{ra} 
\end{figure*}

In order to evaluate our method, we use three types of evaluation measures. First, we consider the automatic natural language generation (NLG) metrics (BLEU \cite{Papineni2001BleuAM}, METEOR \cite{Banerjee2005METEORAA}, ROUGE-L \cite{Lin2004ROUGEAP}, CIDER \cite{Vedantam2014CIDErCI}, SPICE \cite{Anderson2016SPICESP} and BERTScore \cite{Zhang2020BERTScoreET}); all scores are computed with the publicly available code\footnote{https://github.com/tylin/coco-caption}. Second, we use human evaluation as done in previous works. The process of human evaluation is identical to \cite{Kayser2021eViLAD} for VQA-X and e-SNLI-VE, and identical to \cite{Park2018MultimodalEJ} for ACT-X. We still provide full details about the human evaluation process in the Appendix. One drawback of automatic NLG measures is that they do not always reflect the truthfulness and correctness of the explanations, since explanations can come in different forms as well as be very generic and data-biased. Taking VQA as an example, the data-biasness problem refers to models that can correctly answer a question without considering the image information, which is achieved by learning statistics and correlations in the training data. For example, their answer to the question “What is the color of the grass?” is always “green”. Or “What is the color of the banana?” is always
”yellow”. When models are faced with a difficult learning problem, they prefer to learn from the statistical information that links the question with the most occurring answer, completely ignoring the image. By simply memorizing biases in the training data, they exhibit acceptable performance on the test set. As a result, when the model is faced with questions with different image information (e.g. gray grass rather than green grass, or a ripe black banana rather than a yellow banana), they show degraded performance, which also means they are not trustable. The best way to evaluate this phenomenon is conducting human evaluation. However, human evaluation is an expensive and tedious process. To this, automatic measures that better reflect the correctness, context, reasoning, semantic meaning and the degree of biasness of the generated explanations are needed. We therefore propose two new automatic evaluation measures: (1) Explain-Predict, and (2) Retrieval-based attack. We elaborate on these measures below. 

\paragraph{Explain-Predict:} This paradigm is firstly introduced in \cite{Camburu2018eSNLINL}. While \cite{Camburu2018eSNLINL} uses it as the main model, we use it as an evaluation framework. In this evaluation framework, we measure how good the explanation justifies the answer. We input the question and the generated explanation (without the answer) to a language representation model, and aim to predict the answer. This measure gives us a degree on the correlation between the explanation and the answer. If the explanation correctly justifies the answer, we assume the language representation model is able to predict it. 

\begin{figure}
    \centering
    \includegraphics[width=0.5\textwidth]{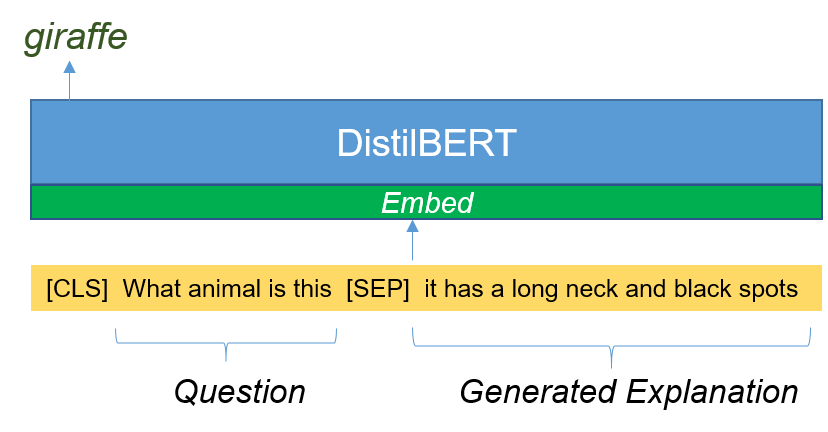}
    \caption{The explain-predict evaluation framework.}
    \label{ep}
\end{figure}

We choose DistilBERT~\cite{Sanh2019DistilBERTAD} to be our language representation model. An example is given in Figure \ref{ep} where the input question is: "what animal is this?" and the generated explanation is "it has a long neck and black spots". Note that we append a CLS token at the beginning of the sequence (as in BERT \cite{Devlin2019BERTPO}) and a SEP token in between the question and the generated explanation, in order to distinguish between them. We take the CLS token output as input to a classification layer over all the answers. In this example, the model predicts "giraffe" as an answer, which correctly justifies the generated explanation. 

\paragraph{Retrieval-based Attack:} In this evaluation framework, we measure the degree our model is susceptible to correlations and bias in the dataset by attacking it with similar inputs. For vision-language tasks, we measure the degree of biasness to both text and images. We will limit our explanation to the biasness to text, as the biasness to images is equivalent. To measure the biasness to text, we consider a text as a query to an image-text retrieval model, and retrieve $K$ similar images close to the text query in the embedding space. For example, the text query could be "is this a healthy meal?" and the retrieved images would be images of different types of meals. Here, we would like to observe the following: given the same question, would the generated explanation always be the same for the different retrieved images? We assume that the same explanations are not due to model reasoning, but rather than correlations and bias in the dataset. In this case we would expect a high cosine intra-distance (distance between the generated explanations of all the retrieved elements). After retrieving the images, we supply our NLX-GPT model with the fixed text query (question) and vary the images (retrieved elements) to generate $K$ different explanations. Given a query, let $\mathcal{G}=\left\{g_{1} \ldots g_{K}\right\}$ be the set of generated explanations for all the $K$ retrieved elements. We feed each $g_k\in \mathcal{G}$ into a language representation model to get its encoded vector representation. By putting together all encoded representations, we obtain a matrix $V\in \mathbb{R}^{K \times d}$, where $d$ is the dimension of the encoded representation. We first perform L2-normalization on each row of $V$; that is, $\bar{v_{k}}=v_{k} /\|v_{k}\|_{2}$, $\forall k=1,\dots, K$. We then compute the gram matrix $\bar{V}\bar{V}^{T}$ of the normalized matrix $\bar{V}$ to find the average cosine distance between each sample with all the other samples as: $s_{\text{avg}}=\frac{1}{K} \sum_{i \in \mathcal{U}} \left[\bar{V} \bar{V}^{T}\right]_i$, and $\mathcal{U}$ is the set of entries in the upper triangular part of $\bar{V} \bar{V}^{T}$ (see green part of the matrix in Figure~\ref{ra}). Note that the negative distances are clamped to 0. Thus, $s_{\text{avg}}$ represents the average intra-distance between the generated explanations of the retrieved elements. The lower the distance is, the lower the bias will be. Therefore, a lower distance is better. We choose Sentence-BERT \cite{reimers-2019-sentence-bert} as the language representation model, a BERT model fine-tuned to contrast between sentence pairs. We use CLIP \cite{Radford2021LearningTV} as the retrieval model. Note that this evaluation framework requires no ground-truth labels, which is advantageous. For vision tasks, since there is no question or hypothesis involved, we utilize an image-retrieval model (i.e., the image part of CLIP~\cite{Radford2021LearningTV}) to retrieve similar images to a given image query, and the remaining process is the same. Figure~\ref{ra} depicts the retrieval-based attack evaluation framework. 

\section{Experiments and Results}
\label{sec:ExperimentalResults}
We mainly experiment with 3 different vision and vision-language NLE datasets: VQA-X \cite{Park2018MultimodalEJ}, ACT-X \cite{Park2018MultimodalEJ} and e-SNLI-VE\cite{Kayser2021eViLAD}. We also experiment with the VCR \cite{Zellers2019FromRT} dataset by re-formulating it as a text generation task. Since the setup of VCR is different from the previously mentioned NLE datasets, we exclude it from the main results and refer readers to the Appendix. VQA-X is a vision-language NLE dataset which extends the Visual Question Answering (VQA v2) dataset \cite{Agrawal2015VQAVQ} and provides explanations for the answers. The images are obtained from the COCO dataset \cite{Lin2014MicrosoftCC}. It contains 33K QA pairs (28K images). ACT-X is a vision NLE dataset which extends the activity recognition dataset \cite{Andriluka20142DHP} and provides explanations for the activity answers. It contains 18K images. Finally, e-SNLI-VE~\cite{Kayser2021eViLAD} is a recently introduced vision-language NLE dataset which provides explanations for the visual entailment prediction task (the task of predicting whether an image and hypothesis \textit{entail, contradict} or are \text{neutral} to each other), and mainly corrects \cite{Do2020eSNLIVE20CV}. The images are obtained from Flickr30k \cite{Plummer2015Flickr30kEC}. It contains over 430K examples. There are currently two different ways previous works evaluate NLE with automatic NLG metrics. The first is evaluating all the explanations in the dataset, regardless of whether the predicted answer for the explanation is true or false. This has been used in \cite{Park2018MultimodalEJ, Wu2019FaithfulME}. We refer to this variant as "unfiltered". The second way is to only consider the explanations for which the predicted answer is correct. This assumes that an explanation is wrong if it justifies a wrong answer, and is therefore not considered in the evaluation. This has been used in \cite{Kayser2021eViLAD}. We refer to this variant as "filtered". We evaluate our method on both variants. 

\begin{table}[]
\centering
\caption{Explain-Predict scores. GT indicates that the ground-truth explanations are fed. Scores are in \%}
\begin{tabular}{|c|cc|}
\hline
\textbf{Dataset} & \textbf{GT} & \textbf{NLX-GPT} \\ \hline
VQA-X            & 86.35       & 77.82           \\ 
e-SNLI-VE         & 93.10       & 73.43           \\ 
ACT-X            & 65.09       & 48.14           \\ \hline
\end{tabular}
\end{table}

\begin{table}[]
\centering
\caption{Retrieval-based attack scores. K indicates how many elements we retrieve for a given query. Lower is better.}
\begin{tabular}{|c|cccccc|}
\hline
         & \multicolumn{3}{c|}{Image Biasness}                        & \multicolumn{3}{c|}{Text Biasness}        \\ \hline
         & \multicolumn{1}{c|}{\textbf{K@5}}    & \multicolumn{2}{c|}{\textbf{K@10}}   & \multicolumn{2}{c|}{\textbf{K@5}}   & \textbf{K@10}         \\ \hline
VQA-X    & \multicolumn{1}{c|}{44.56}  & \multicolumn{2}{c|}{44.99}  & \multicolumn{2}{c|}{47.84} & 46.26        \\ 
e-SNLI-VE & \multicolumn{1}{c|}{39.79}  & \multicolumn{2}{c|}{37.95}  & \multicolumn{2}{c|}{67.16} & 65.69        \\ \hline
         & \multicolumn{6}{c|}{Image Biasness}                                                                   \\ \hline
         & \multicolumn{2}{c|}{\textbf{K@5}}                    & \multicolumn{2}{c|}{\textbf{K@10}}  & \multicolumn{2}{c|}{\textbf{K@15}}  \\ \hline
ACT-X    & \multicolumn{2}{c|}{63.78}                  & \multicolumn{2}{c|}{55.08} & \multicolumn{2}{c|}{48.19} \\ \hline
\end{tabular}
\end{table}

\subsection{Quantitative Analysis}
Table 2 shows our results on the "unfiltered" variant. As seen, our model outperforms all previous works on both VQA-X and ACT-X. Table 3 shows our results on the "filtered" variant. Our model also outperforms all previous models on most of the scores on both VQA-X and e-SNLI-VE, while being much more memory efficient, a lot more faster, and requiring no regional and strong multi-modal features. We also show our task performance accuracy scores. For VQA-X, we consider an answer to be correct if it is included in the set of all possible ground-truth answers. In Table 4, we report our results on our proposed explain-predict evaluation framework. We also show the results when feeding the ground-truth explanations (column indicated by "GT"). This gives a measure on the top-performing score. In Table 5, we show our results on our proposed retrieval-based attack framework. These are computed by averaging over all the intra-distances scores for all the test data (images or text) in the respective dataset (lower is better). We report scores when we retrieve $K=5,10,15$ elements for a given query. 

\subsection{Qualitative Analysis}
\begin{figure}
    \centering
    \includegraphics[width=0.5\textwidth]{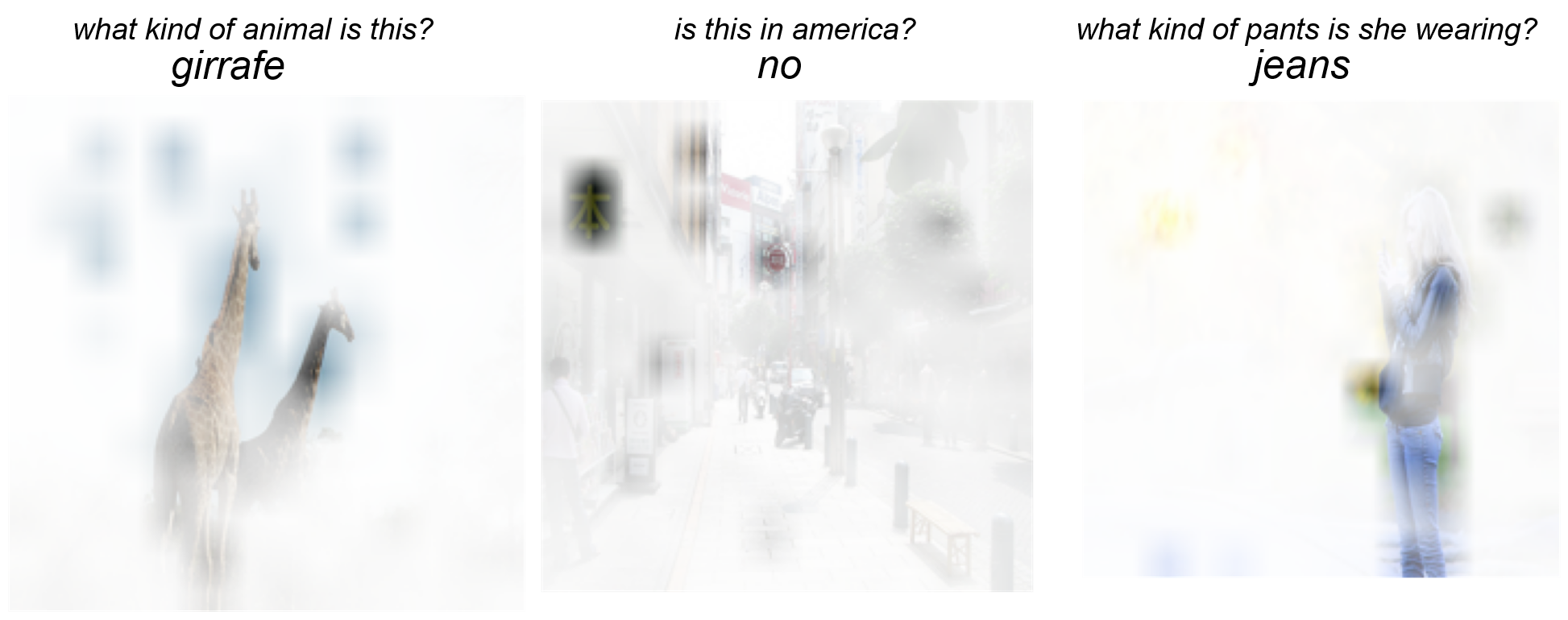}
    \caption{The attention maps for the predicted answer}
    \label{att_viz}
\end{figure}

\begin{figure}
    \centering
    \includegraphics[width=0.5\textwidth]{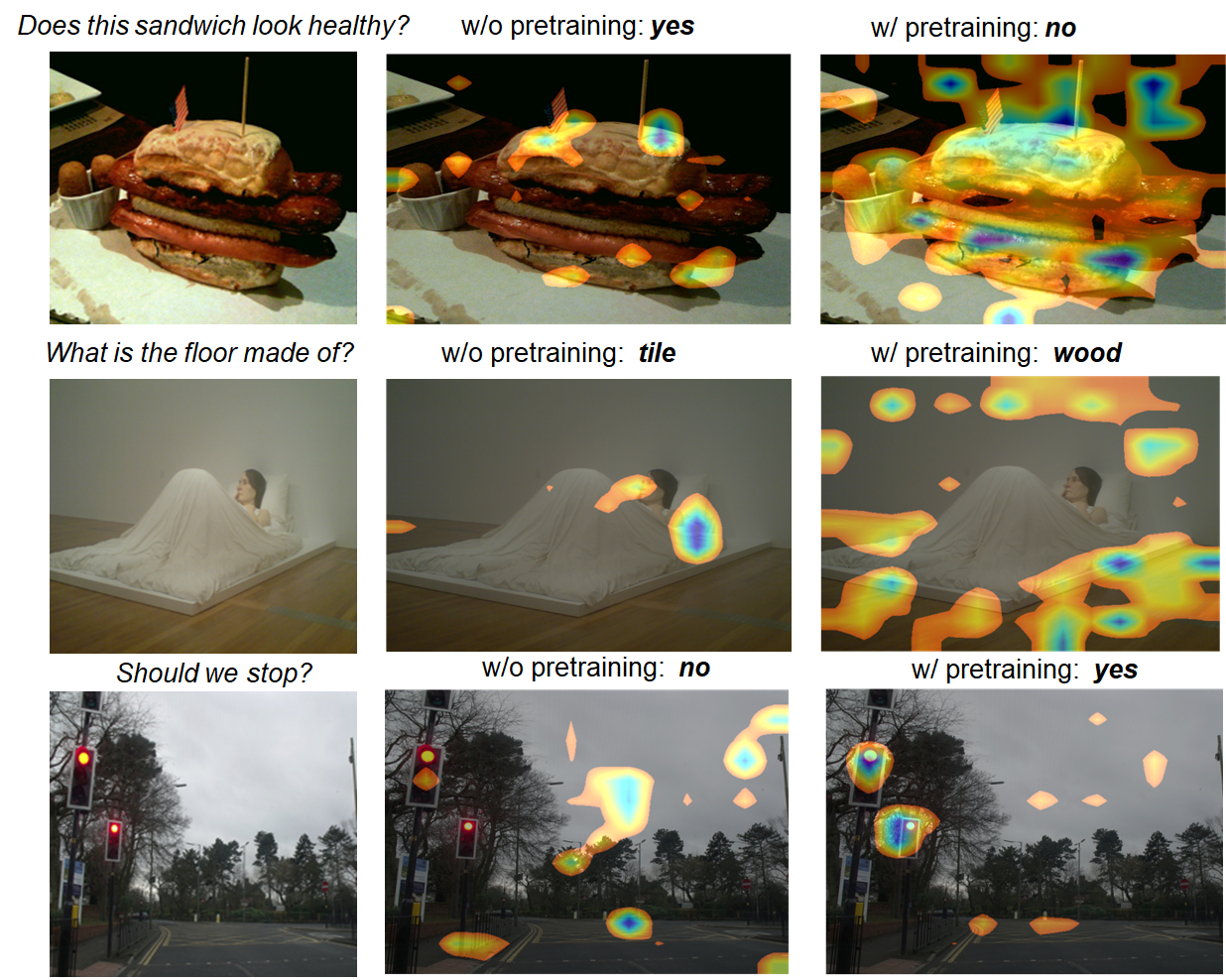}
    \caption{The attention maps for the predicted answer without pretraining (middle) and with pretraining (right)}
    \label{att_viz2}
\end{figure}

\begin{figure*}
    \centering
    \includegraphics[width=0.97\textwidth]{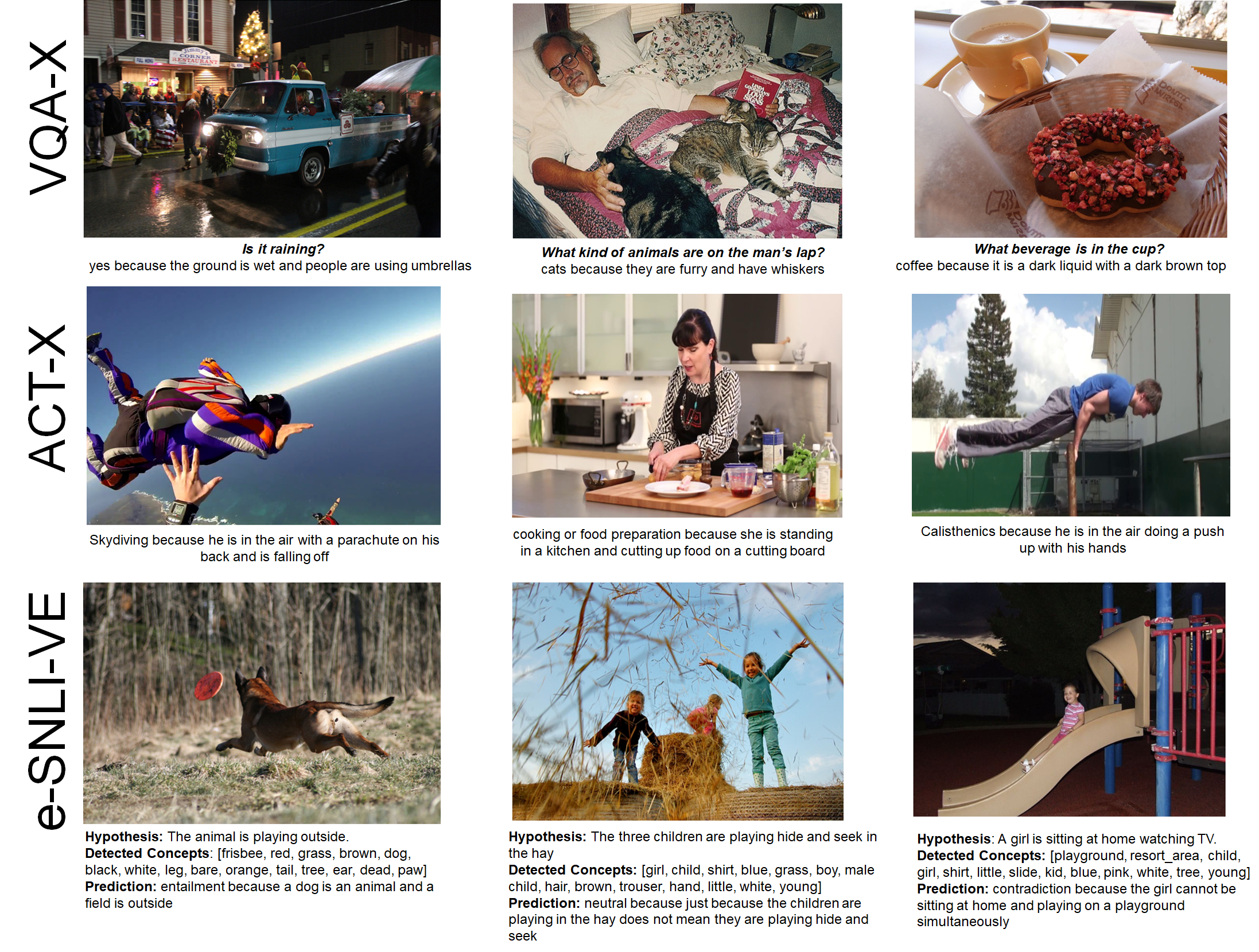}
    \caption{Some qualitative examples from our model on the three tasks}
    \label{examples}
\end{figure*}

In Figure \ref{att_viz} we visualize the attention maps for the predicted answer, which is modeled as a text prediction task along with the explanation. In Figure \ref{att_viz2}, we visualize the attention map for the predicted answer without and with model pretraining. This clearly shows how model pretraining helps the model to reason and visual-ground the correct answer on which the explanation will be conditioned on. In Figure \ref{examples}, we show some qualitative results from our model on all three tasks. We refer to the Appendix for more qualitative examples, failure cases and retrieval-based attack examples. 

\subsection{Ablation Studies}
We ablate 2 different aspects of our model: First, we wish to observe how much performance does pretraining (for VQA-X and ACT-X) or concept detection (for e-SNLI-VE) add to the overall performance. The last two rows in Tables 2 and 3 demonstrate this effect. Second, we ablate different image encoders. Particularly, we experiment with the vision transformer \cite{Dosovitskiy2020AnII}, ResNet-101 \cite{He2016DeepRL}, DeiT \cite{Touvron2021TrainingDI} and bottom-up features \cite{Anderson2018BottomUpAT}. Table 6 demonstrates our results on these image encoders. The best performance is obtained by the CLIP vision transformer \cite{Radford2021LearningTV}. Different from \cite{Shen2021HowMC}, we find the CLIP vision transformer to outperform the CLIP ResNet-101. We also show the superiority of our model even with a ResNet-101 pretrained on ImageNet-1K \cite{Krizhevsky2012ImageNetCW} or with bottom-up features \cite{Anderson2018BottomUpAT}. Note that the scores in Table 6 are \textit{without} image-caption pretraining. 

\begin{table}
\centering
\caption{Ablation studies on different image encoders. The scores are \textit{filtered} as in Table 3. Also note that the scores are \textit{without} image captioning pretraining. $^{\dagger}$ indicates that we use the CLIP model \cite{Radford2019LanguageMA} weights. $^{\star}$ indicates the model is pretrained on ImageNet-1K. ↑ means the vision backbone is fine-tuned. Otherwise, the vision backbone is fixed.}
\scalebox{0.9}{
\begin{tabular}{|c|c|c|c|c|}
\hline
Image   Backbone & BLEU-4 & METEOR & CIDER & ROUGE \\ \hline
ViT$^{\dagger}$             & 28.1   & 22.6   & 108.5 & 50.9  \\ 
ResNet-101$^{\dagger}$      & 26.7   & 22.1   & 102.3 & 50.6  \\ 
DeiT$^{\star}$             & 26.4   & 21.8   & 97.6  & 49.7  \\ 
ResNet-101$^{\star}$       & 22.4   & 19.5   & 81.6  & 45.7  \\ 
ResNet-101$^{\star}$↑     & 25.2   & 20.5   & 95.2  & 47.3  \\ 
BU-feats        & 26.5   & 22.5   & 101.1  & 50.1  \\   \hline
\end{tabular}
}
\end{table}

\section{Limitations}
\textit{There is no free lunch --} Although our model brings many advantages, it still has some limitations. It performs slightly worse than \cite{Kayser2021eViLAD} on the e-SNLI-VE task for three automatic NLG metrics: METEOR, SPICE and BERTScore. These metrics have shown high correlation with human judgment in \cite{Kayser2021eViLAD}. Our model thus favors N-gram (such as BLEU, ROUGE) and human consensus (such as CIDER) metrics on the e-SNLI-VE task and gives more weight to those metrics rather than the others. We also observe that other NLE models on the e-SNLI-VE task give more weight to either the former or the later group of metrics, but not both. This is also observed in \cite{Kayser2021eViLAD} where the BLEU-N, R-L and CIDER scores are lower than ours. 

\section{Conclusion and Future Directions}
We proposed NLX-GPT, a model which can simultaneously predict an answer and explain it by formulating the answer prediction as a text generation task along with the explanation. We also proposed two new evaluation frameworks to better evaluate the generated explanations. In the future, we would like to take advantage of powerful language understanding models through techniques such as distillation, or even by leveraging NLE datasets aimed for NLP tasks, which are much more diverse and of large-scale, and thus can greatly benefit NLE vision-language models. We also expect to see self-critical sequence training \cite{Rennie2017SelfCriticalST} or its better variants \cite{Luo2020ABV} incorporated. 

\paragraph{Acknowledgement:} This research has been supported by the Research Foundation - Flanders (FWO) under the Project G0A4720N.

{\small
\bibliographystyle{ieee_fullname}
\bibliography{egbib}
}

\clearpage

\setcounter{section}{0}
\setcounter{figure}{0} 
\setcounter{table}{0}

{\Large \textbf{Supplementary Material}}
\newline

\section{Implementation Details}
In this section, we provide implementation details for pretraining, concept detection and finetuning. Please note that for all three stages, the vision backbone is frozen and not fine-tuned at any time. 

\subsection{Pretraining}
We choose 4 publicly available image-caption datasets for pre-training: MSCOCO \cite{Lin2014MicrosoftCC}, Visual Genome Region Descriptions \cite{Krishna2016VisualGC}, Flickr30K \cite{Plummer2015Flickr30kEC} and Image-Paragraph Stanford dataset \cite{Krause2017AHA}. The region descriptions of Visual Genome are of large-scale ($	\sim$5M). However, they are short compared to COCO and Flickr30K, and each region description is associated to a small part of the image. We therefore combine the per-image region descriptions to form a paragraph, which acts as the caption associated to the image. The maximum length of the caption is set to 70. There are also other image-caption datasets that can be used for pretraining, such as Conceptual Captions of $\sim$3M pairs \cite{Sharma2018ConceptualCA} as well as SBU of $\sim$1M pairs \cite{Ordonez2011Im2TextDI}. We leave these to future works since the results with the 4 datasets we mentioned already give satisfactory results when fine-tuned on the Natural Language Explanations (NLE) downstream task. We initialize the model with the Distilled GPT-2 weights\footnote{https://github.com/huggingface/transformers}. The model is trained with the ADAM optimizer \cite{Kingma2015AdamAM} with a batch size of 768 and a learning rate of 1e-4 which is linearly decayed to 0 over the total number of training steps. We evaluate the pre-trained model performance on the "Karpathy" test split \cite{Karpathy2017DeepVA} of COCO Captions \cite{Lin2014MicrosoftCC}.

\subsection{Concept Detection}
Let $H, W, P, Y$ be the height, width, patch size and total number of patches of the image, respectively. In order to predict image concepts, the output representation of the vision backbone for all image patches is utilized. Let the output of the vision backbone be $X \in \mathbb{R}^{Y \times d}$ where $d$ is the output dimension. In the case of ViT, we do not utilize the {\fontfamily{qcr}\selectfont CLS} reduced representation since it is mainly optimized for other objectives such as image classification or contrastive learning, and may not be optimal for concept learning which requires a much broader view of all the image patches. Therefore, we learn an attention reduced representation of all the $Y$ image patches. We first feed $X$ into 2 linear layers with a ReLU activation function in-between, followed by a residual \cite{He2016DeepRL} and layer normalization layer \cite{Ba2016LayerN} to get an output  $V \in \mathbb{R}^{Y \times d_{k}}$. We utilize an attention-summary layer implemented as $\mathbf{\textbf{s}}=\sum_{i=1}^{Y} \alpha_{i} \mathbf{x}_{i}$ where $\alpha_{i}=\frac{\exp \left(\mathbf{w}_{x}^{T} \mathbf{v}_{i} \right)}{\sum_{j=1}^{Y} \exp \left(\mathbf{w}_{x}^{T} \mathbf{v}_{j} \right)}$ in order to reduce $X$ into to a single feature vector $\textbf{s}$, where $w \in \mathbb{R}^{d_{k} \times 1}$ are learnable weights.  $\textbf{s}$ is then fed into a classification layer over all the concepts. We train the concept detection head on the Visual Genome 2.8M attributes dataset \cite{Krishna2016VisualGC} with a batch size of 256 using the ADAM optimizer \cite{Kingma2015AdamAM} with a learning rate of 2e-3 which is decayed by a factor of 0.8 every 3 epochs. We use dropout with a keeping probability of 0.9. In order to measure the accuracy, we take the top-K predicted concepts and check how many of them appear in the ground-truth concepts. For each sample in the test set, we count the number of elements in the set ${PR \cap GT}$, where $PR$ indicates the top-K predicted concepts and $GT$ are all the ground-truth concepts for the tested sample. We measure the acuracy at $K = 5, 10$ and $15$. Our resulting accuracy scores are 83.0, 73.61 and 65.99, respectively. 

\subsection{Finetuning}
We finetune our pretrained model on 4 NLE datasets: VQA-X, ACT-X, e-SNLI-VE and VCR (explained in the next section). The input sequence consists of tokens of the (question, answer, explanation) and each token is then fed to an embedding layer to get a representation of the word. In order allow the model to distinguish between the question, answer and explanation, we add the embeddings of the segment ID {\fontfamily{qcr}\selectfont <ques>}, {\fontfamily{qcr}\selectfont <ans>} and {\fontfamily{qcr}\selectfont <exp>} to all question, answer and explanation token embeddings, respectively. We also add a continuous positional embedding for the complete sequence starting from the question up until the explanation. We use the ADAM optimizer \cite{Kingma2015AdamAM} with a learning rate of 1e-5 which is linearly decayed to 0 over the total number of training steps. For models which are finetuned from the pretrained image captioning model (models for VQA-X and ACT-X), we use a batch size of 32. For other models, we initialize them from the Distilled GPT-2 weights and use a batch size of 64. At inference, we use greedy decoding. The maximum sequence length for VQA-X, ACT-X and e-SNLI-VE is set to 40, 30, and 40, respectively. For e-SNLI-VE, the maximum number of concepts fed at the input is 15. 

\begin{table*}[]
\centering
\caption{Filtered Scores on VCR dataset. B, M, R-L, C, S, BS are short for BLEU, METEOR, ROUGE-L, CIDER, SPICE and BERT Score, respectively. Unfiltered Results for B1, B4, M, R-L, C, S, BERTScore are: 18.5, 3.3, 9.0, 19.9, 24.2, 12.4, 77.1}
\begin{tabular}{cccccccccc}
\hline
        & B-1           & B-2           & B-3          & B-4          & M    & R-L           & C             & S    & BS            \\ \hline
PJ-X \cite{Park2018MultimodalEJ}    & 21.8          & 11.0          & 5.9          & 3.4          & 16.4 & 20.5          & 19.0          & 4.5  & 78.4          \\ 
FME \cite{Wu2019FaithfulME}     & 23.0          & 12.5          & 7.2          & 4.4          & \textbf{17.3} & 22.7          & 27.7          & \textbf{24.2} & 79.4          \\ 
RVT \cite{Marasovi2020NaturalLR}     & 18.0          & 10.2          & 6.0          & 3.8          & 11.2 & 21.9          & 30.1          & 11.7 & 78.9          \\ 
QA-only \cite{Kayser2021eViLAD} & 18.0          & 10.2          & 6.0          & 3.8          & 11.2 & 22.0          & 30.6          & 11.6 & 78.9          \\ 
e-UG  \cite{Kayser2021eViLAD}   & 20.7          & 11.6          & 6.9          & 4.3          & 11.8 & 22.5          & 32.7          & 12.6 & 79.0          \\ \hline
NLX-GPT & \textbf{24.7} & \textbf{15.0} & \textbf{9.6} & \textbf{6.6} & 12.2 & \textbf{26.4} & \textbf{46.9} & 18.8 & \textbf{80.3} \\ \hline
\end{tabular}
\end{table*}

\section{VCR Setup and Experiments}

\begin{figure*}
    \centering
    \includegraphics[width=0.95\textwidth]{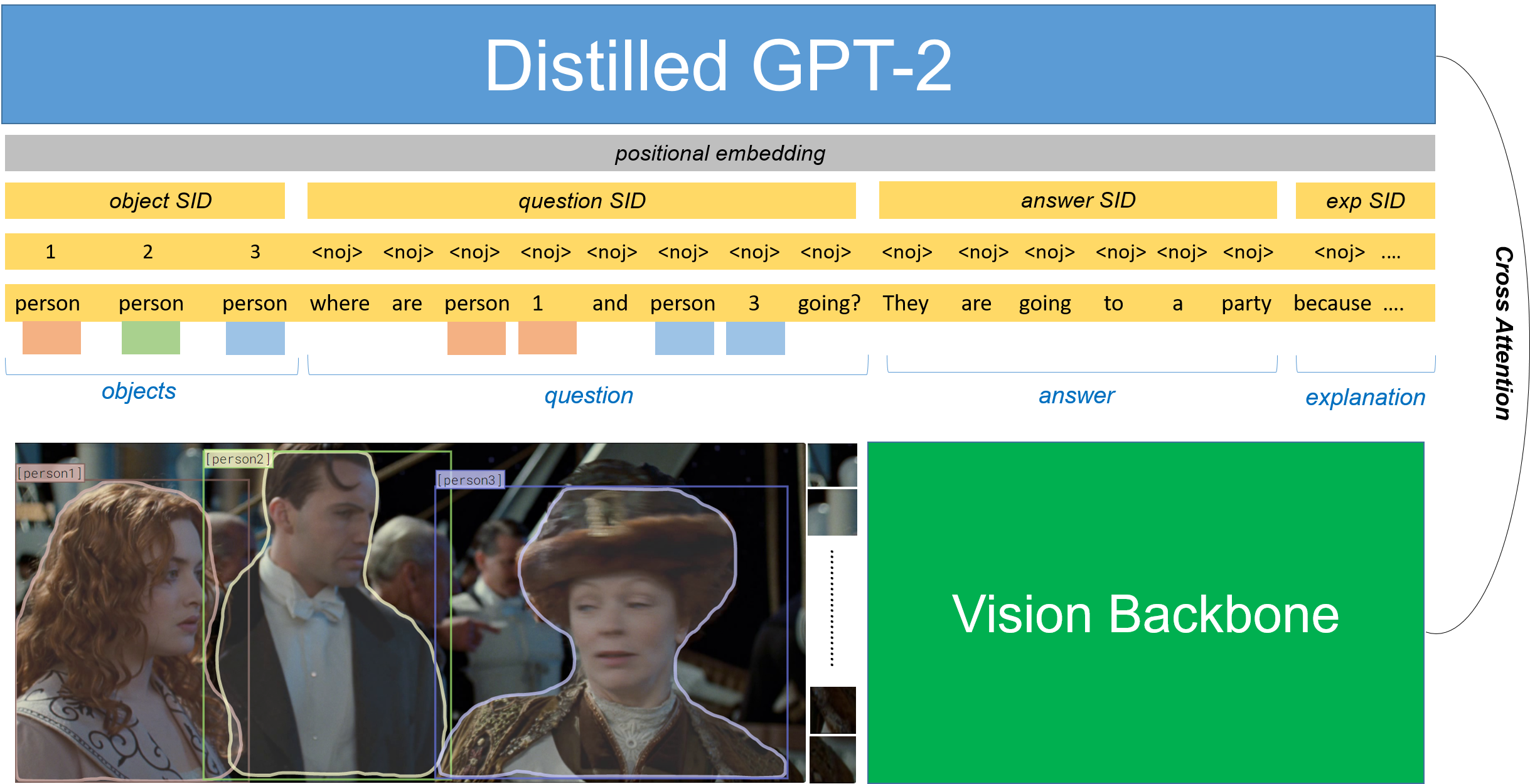}
    \caption{Our proposd NLX-GPT for the VCR Task. The word embedding, object reference number embedding as well as the segment ID embeddings all share the same layer (orange). The red, green and blue squares represent the projections of the bounding box coordinates for person 1,2 and 3, respectively.}
    \label{vcrschema}
\end{figure*}

\begin{figure*}
    \centering
    \includegraphics[width=0.9\textwidth]{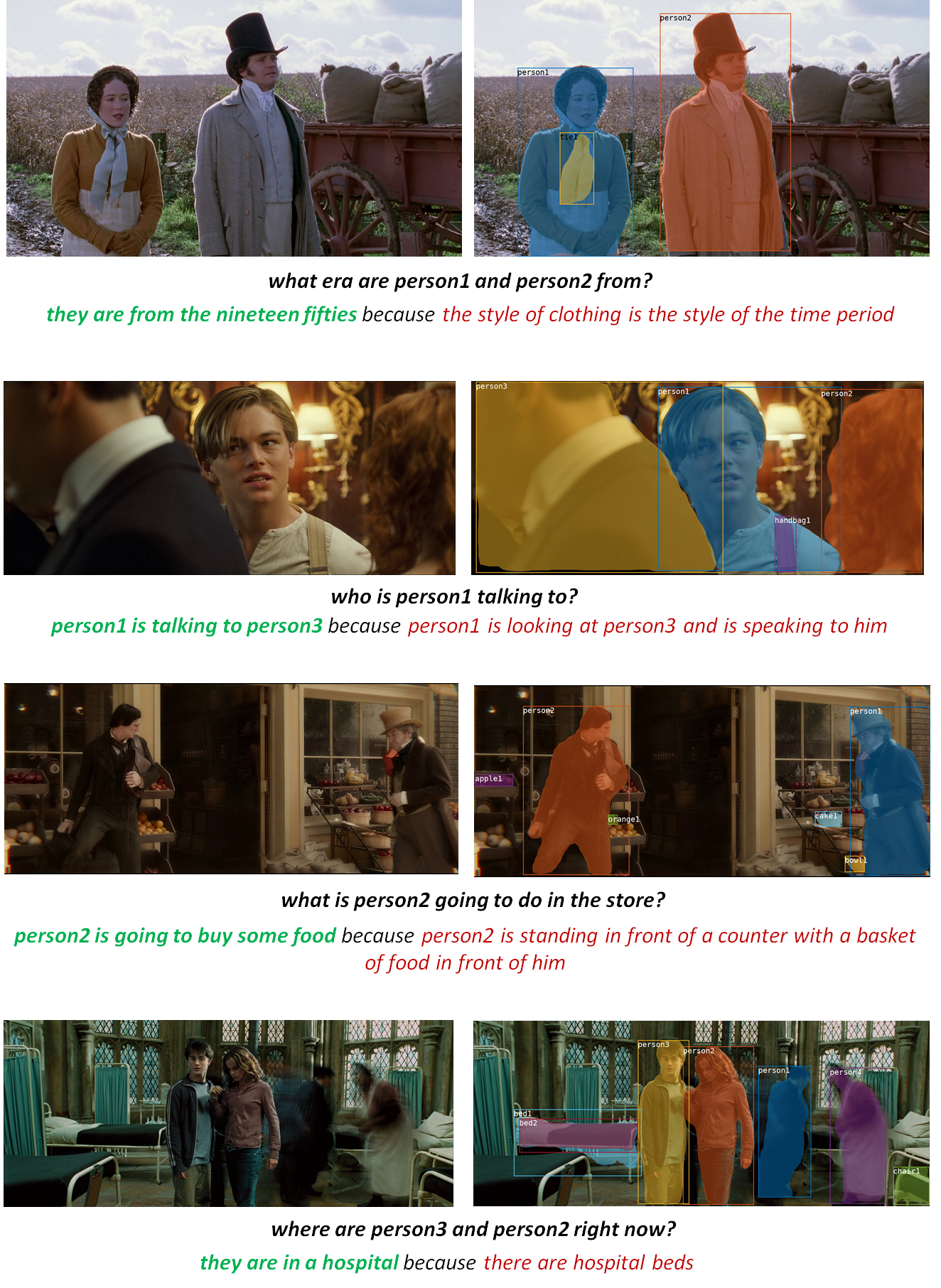}
    \caption{Qualitative examples from our model on the VCR task}
    \label{vcrex}
\end{figure*}

Visual Commensense Reasoning (VCR) \cite{Zellers2019FromRT} is a new task introduced in which a model is given an image, question and a list of objects (regional features and bounding boxes) and is required to select one answer from a set of multiple-choice answers ($Q \rightarrow A$). After that, it is required to select a rationale (explanation) of \textit{why} the answer it has selected is correct, from a set of multiple-choice rationales ($QA \rightarrow R$). The dataset consists of 290K samples of questions, answers and rationales. For the purpose of the NLE task, we follow previous NLE models \cite{Kayser2021eViLAD, Marasovi2020NaturalLR} and reformulate the explanation as a text generation task rather than a multiple-choice answering task. The train, validation and test splits for NLE are 191.6k, 21.3k, and 26.5k, respectively. Different from previous NLE vision-language tasks (VQA-X and e-SNLI-VE) where the input is an image and a question/hypothesis, VCR requires an additional input (detected objects) which can be represented by the regional features. Given the region proposal coordinates, NLE models implementing the re-formulated VCR \cite{Marasovi2020NaturalLR, Kayser2021eViLAD} first extract these regional features by performing Region-of-Interest (ROI) pooling or ROI Align on the output of a Faster R-CNN network \cite{Ren2015FasterRT}. Let a bounding box for a specific object be represented by $x_{1}, y_{1}, x_{2}, y_{2}$ which indicates the top-left and bottom-right coordinates. One approach we could take to represent objects for our NLX-GPT is to also perform ROI pooling on the grid-based vision backbone output. In the case of the vision transformer, we can reshape the output of shape $Y \times D$ to $H^\prime \times W^\prime \times D$, where $H^\prime = H/P$ and $W^\prime = W/P$ and $D$ is the output dimension. After that, we can perform ROI pooling on that reshaped output using the given bounding box coordinates. However, we take a simpler approach. 
We first represent each bounding box with 8 values: $(\frac{x_{1}}{W}, \frac{y_{1}}{H}, \frac{x_{2}}{W}, \frac{y_{2}}{H}, \frac{x_{1}+x_{2}}{2 W}, \frac{y_{1}+y_{2}}{2 H}, \frac{x_{2}-x_{1}}{W}, \frac{y_{2}-y_{1}}{H})$, which are then projected to a high-dimensional representation equal to the dimension of the word and positional embeddings, in order to represent the object positional information. Since the question, answer or explanation may refer to specific detected objects in the image (e.g., \textit{what are \textbf{person1} and \textbf{person3} doing?}, it becomes necessary to encode these objects along with their respective reference number. We therefore input to the Distilled GPT-2 tokens which consist of the objects, question, answer and explanation (as a single sequence). We embed these input tokens with a token embedding layer. We also embed the object reference number (ORN) with the same token embedding layer. We use {\fontfamily{qcr}\selectfont <noj>} (representing "no object") to represent the ORN for the question, answer and explanation tokens. Finally, we add the object positional information, token embeddings, ORN embeddings, positional embedding and segment embeddings together to form the Distilled GPT-2 input. The maximum number of objects is set to 20, and the maximum length of the (question, answer, explanation) sequence is set to 60. Figure \ref{vcrschema} illustrates the complete process. For the purpose of our NLX-GPT, we also formulate the answer prediction as a text generation task. Unlike previous tasks discussed in the main paper (VQA-X, ACT-X and e-SNLI-VE) where the answer consists of one or a maximum of two words, the answer in VCR is usually much longer. It is therefore difficult to expect an identical correspondence between the generated and ground-truth answer. For example, the model may generate an answer: \textit{no, this person does not live in this house} while the ground-truth answer is: \textit{no, this person is a visitor}. In fact, this justifies why the authors of \cite{Zellers2019FromRT} formulated the VCR task as a multiple-choice task. At evaluation, only the test samples for which the predicted answer is correct are allowed to proceed to the second stage of providing the rationale ($QA \rightarrow R$), and thus test samples with wrong predicted answers should be filtered. To overcome the difficulty, we measure the context and semantic meaning of our predicted answers through the BERTScore \cite{Zhang2020BERTScoreET} metric. We thus consider a predicted answer to be correct if its BERTScore referenced with its corresponding ground-truth is higher than or equal to 0.92. Table 1 shows our filtered scores and Figure \ref{vcrex} shows qualitative examples. 

\begin{figure*}
    \centering
    \includegraphics[width=\textwidth]{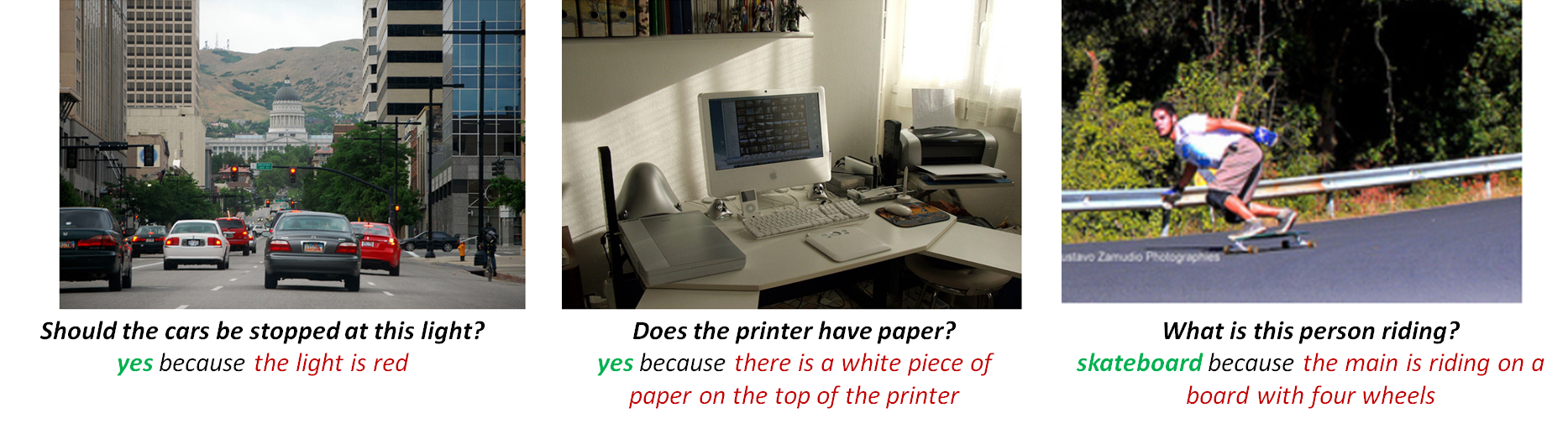}
    \caption{More qualitative examples from our model on the VQA-X task}
    \label{vqaxex}
\end{figure*}

\begin{figure*}
    \centering
    \includegraphics[width=\textwidth]{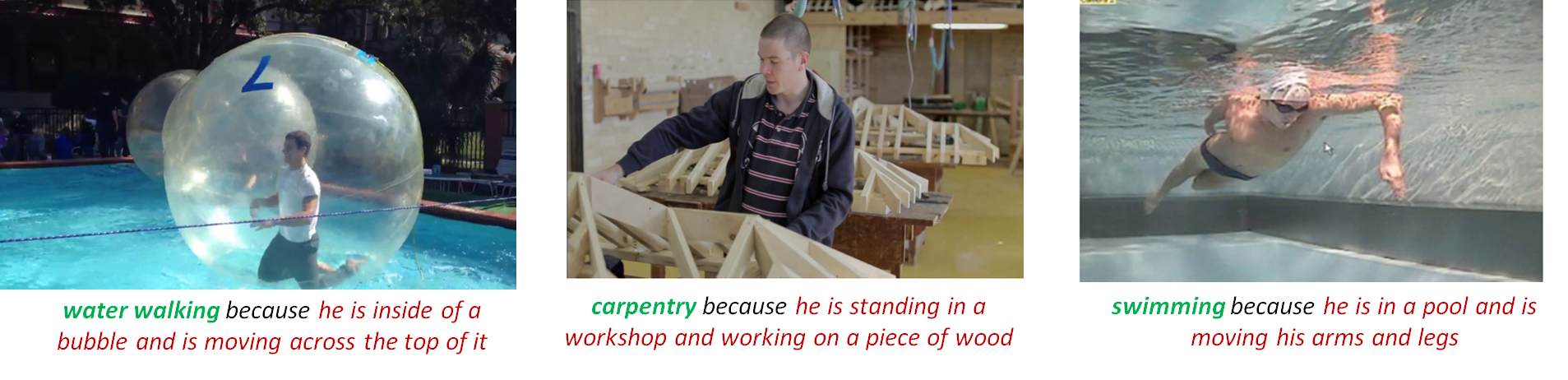}
    \caption{More qualitative examples from our model on the ACT-X task}
    \label{actxex}
\end{figure*}

\begin{figure*}
    \centering
    \includegraphics[width=\textwidth]{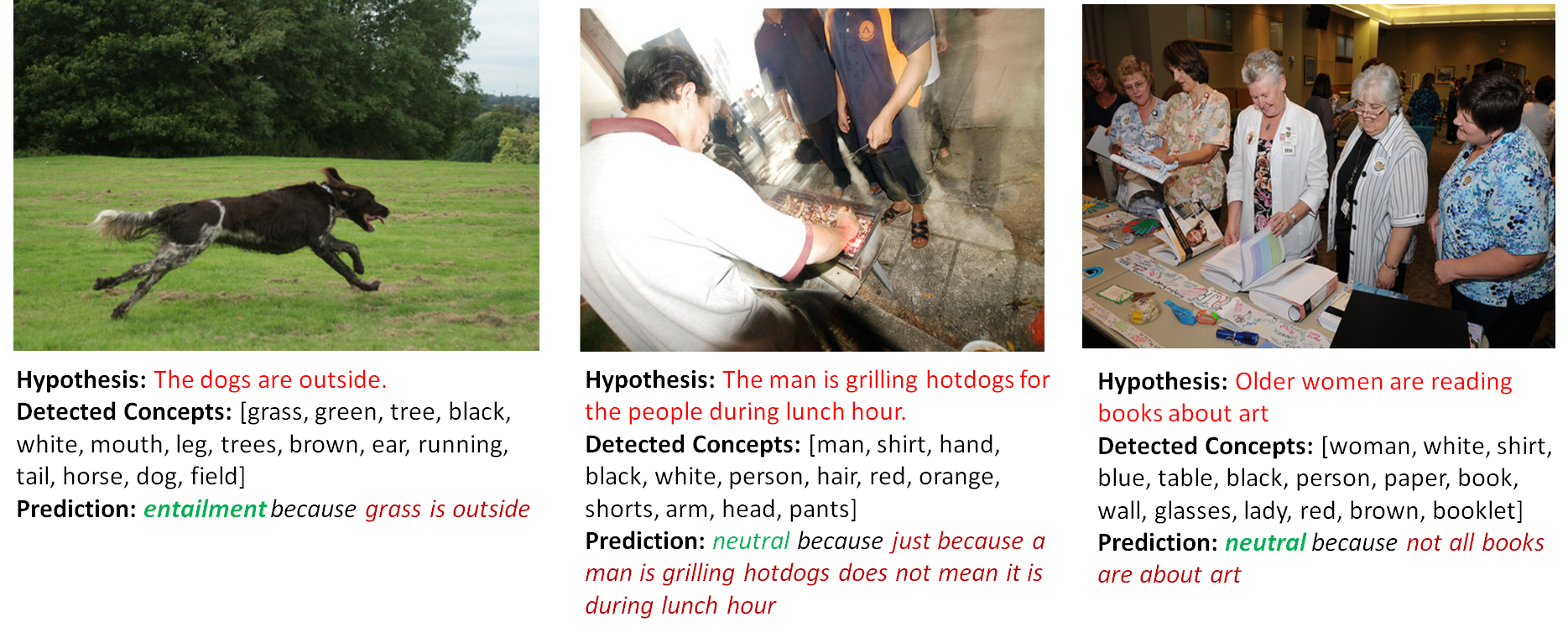}
    \caption{More qualitative examples from our model on the e-SNLI-VE task}
    \label{esnliveex}
\end{figure*}

\begin{figure*}
    \centering
    \includegraphics[width=\textwidth]{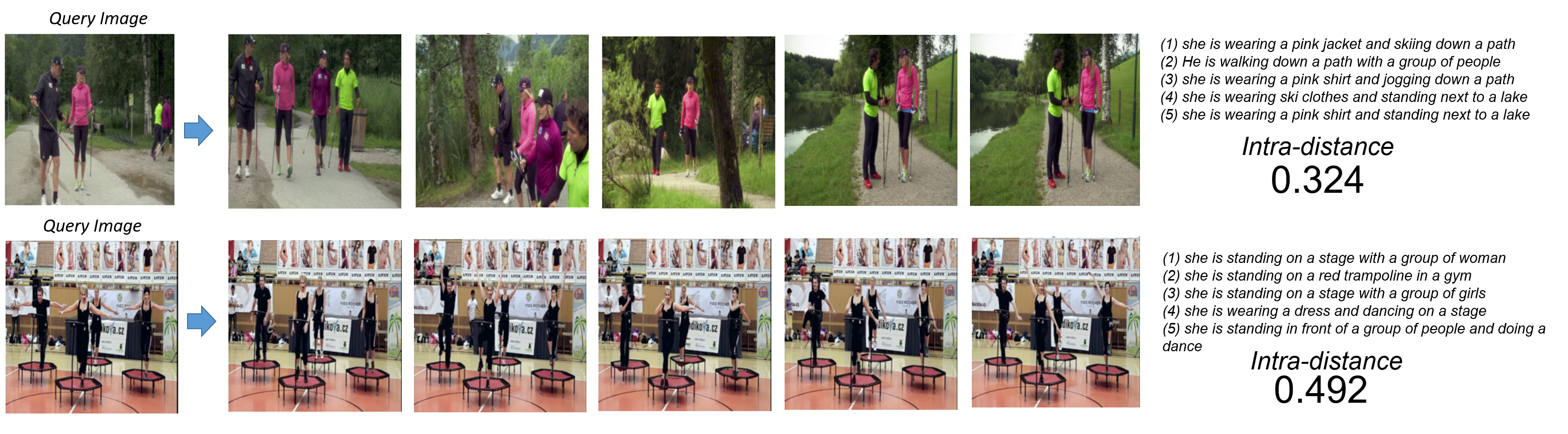}
    \caption{Retrieval-based attack evaluation results visually for two test samples from the ACT-X dataset}
    \label{ra_examples}
\end{figure*}

\begin{figure*}
    \centering
    \includegraphics[width=\textwidth]{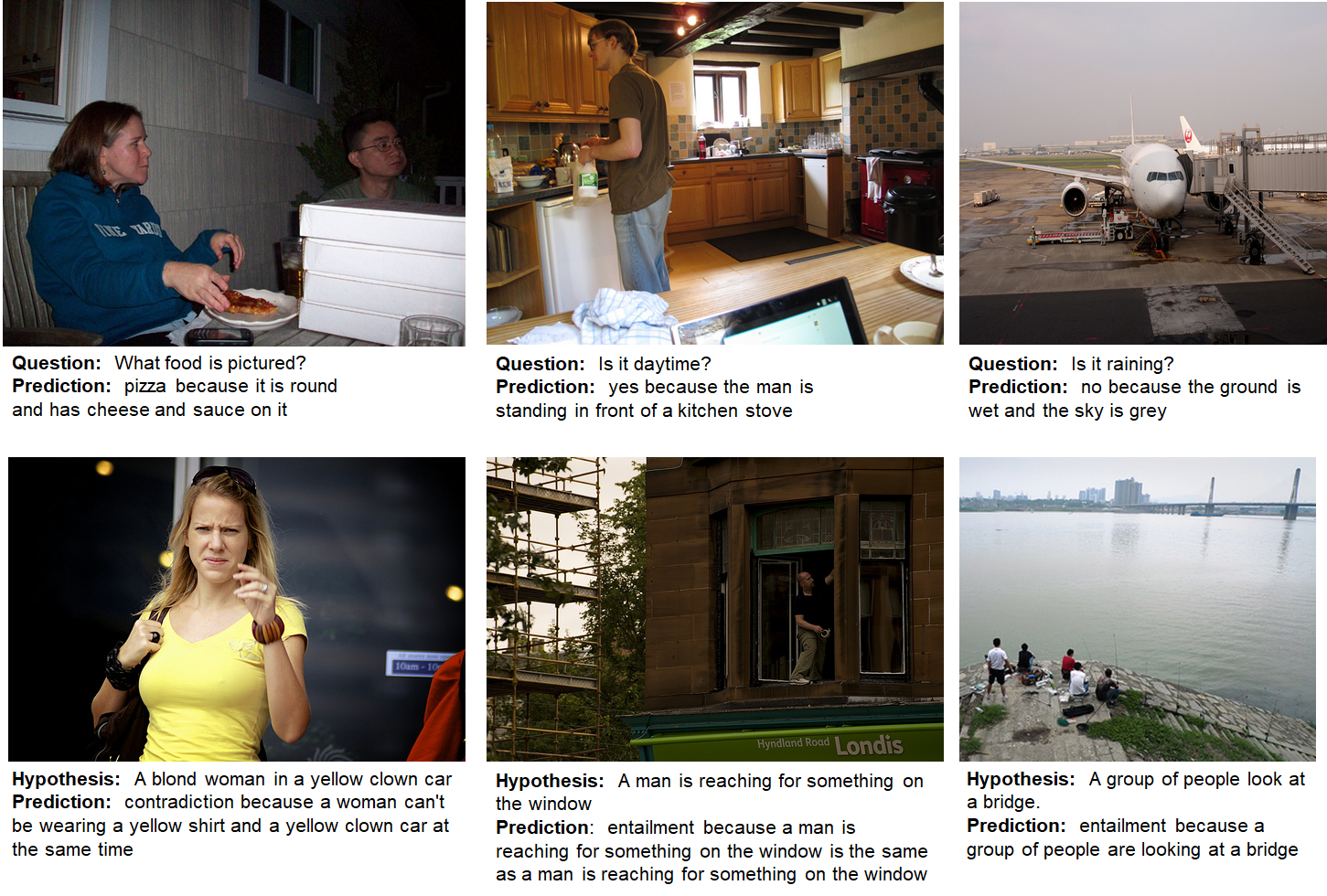}
    \caption{Failure cases on the VQA-X and e-SNLI-VE tasks}
    \label{failure}
\end{figure*}

\section{Explain-Predict Implementation Details}
The explain-predict evaluation framework is trained on the ground-truth explanations. That is, we feed the question and ground-truth explanation to the Distilled-BERT model during training. We initilze the model with the Distilled-BERT weights\footnote{https://github.com/huggingface/transformers}. The model is trained with the ADAM optimizer with a batch size of 16 and a learning rate of 2e-5 which is linearly decayed to 0 over the total number of training steps. For VQA-X and ACT-X, we train a multi-label classifier with soft targets over the ground-truth answers using binary cross-entropy loss. For e-SNLI-VE, we train a a classifier with the hard targets over the ground-truth answers using cross-entropy loss. We measure the accuracy as the evaluation criteria. It is important to note that for VQA-X, we find a total of 37 test samples for which their answers are never seen in the training set. For other NLE models \cite{Park2018MultimodalEJ, Wu2019FaithfulME, Kayser2021eViLAD}, that is not a problem since they employ a pretrained VQA model (trained on the full VQAv2 dataset). However, our NLX-GPT is trained from scratch and only on the VQA-X dataset (which is much smaller than VQAv2). We therefore exclude these 37 samples from the explain-predict accuracy calculation. 

\section{More Qualitative Examples}
We include more qualitative examples from VQA-X, ACT-X and e-SNLI-VE in Figures \ref{vqaxex}, \ref{actxex}, and \ref{esnliveex}. Figure \ref{ra_examples} depicts the retrieval-based attack evaluation results visually for two test samples from the ACT-X dataset when $K=5$. As shown, very similar images have a low intra-distance, and thus our model has a low susceptibility to correlations and bias in the dataset. 

\section{Human Evaluation Process}
For VQA-X and e-SNLI-VE, the human evaluation process is identical to \cite{Kayser2021eViLAD}. We randomly select 300 test samples with correctly predicted answers. The evaluation is performed by 3 different annotaters and the results are then averaged. The annotaters mainly have to select one out of 4 choices (yes, weak yes, weak no, and no) as a response for whether the explanation justifies the answer. The 4 scores are numerically mapped to 1, 2/3 , 1/3 , and 0, respectively. The numerical scores are then averaged among all test samples to get a final score. For ACT-X, we follow the main procedure introduced in the paper \cite{Park2018MultimodalEJ}. 300 test samples with correctly predicted answers are randomly chosen, and a human annotater is asked to determine whether a generated explanation is superior to, inferior to, or equal to the ground truth explanation. The percentage of the generated explanations which are equal to or superior to the ground truth explanations are reported. 

\section{Failure Cases}
We include failure cases of our model in Figure \ref{failure}. We realize that in some cases, the answer is predicted wrong but the explanation is correct or vice-versa. For cases where the explanation is wrong but the answer is correct, we hypothesize the problem to be in the equal weighing of all words to be generated (including the answer), which treats the answer as all other words. To alleviate this problem, we tried several solutions. One solution is to assign more weight to any word in the answer vocabulary during the loss calculation, so that wrong wrongly predicted answers incur more loss. Unfortunately, none of the solutions we tried gave positive effects. For e-SNLI-VE, we observe two major failure cases. Firstly, the model justifies the prediction by simply repeating the hypothesis. Secondly, the model is sometimes biased towards the tone the human explainer justifies the prediction, as many samples in the dataset are of the form: (\textit{.....is the same as.....}) or (\textit{just because.....does not mean.....}). 

\section{Removing a VL-model is advantageous}
In the paper we give two intuitions behind this: In short, 1) it eliminates the high memory requirements of the VL-model and reduces the inference time. 2) it eliminates the independence and dissociation of the VL-model and explanation model, in the sense that the explanation is intrinsic, internally affiliated and connected to the reasoning process made to predict the answer. There are other reasons as well: Training the task and explanation jointly in one model allows us to have faithful explanations, which is what FME\cite{Wu2019FaithfulME} discusses, however in a much simpler and model-intrinsic way, and without any external operations. Also, if we had a separate VL-model and a separate explanation model, finetuning the VL-model along with the explanation model is advantageous, but a difficult step (in most works, this step is avoided) due to the extra memory requirements to fine-tune the VL-model as well as the careful and correct consideration of hyperparameters and optimization procedure required. In our work, this is completely avoided since both are jointly trained in one model.

\end{document}